\def\year{2015}
\documentclass[letterpaper]{article}
\usepackage{aaai}
\usepackage{times}
\usepackage{helvet}
\usepackage{courier}
\usepackage{color}
\usepackage[dvips,final]{graphicx}
\usepackage{amsmath}
\usepackage{amsthm}
\usepackage{algorithm}
\usepackage{algorithmic}
\usepackage{amsfonts}
\usepackage{subfigure}
\usepackage{url}
\usepackage{collcell}
\usepackage{datatool}
\usepackage{bm}
\usepackage{bbm}
\usepackage{multirow}
\usepackage{multicol}
\theoremstyle{plain}
\begin{document}
%
\title{Learning Deep $\ell_0$ Encoders}

\author{Zhangyang Wang \dag , Qing Ling \ddag, Thomas S. Huang \dag \\
\dag Beckman Institute, University of Illinois at Urbana-Champaign, Urbana, IL 61801, USA\\
\ddag Department of Automation, University of Science and Technology of China, Hefei, 230027, China
}

\maketitle
\begin{abstract}
\begin{quote}
Despite its nonconvex nature, $\ell_0$ sparse approximation is desirable in many theoretical and application cases. We study the $\ell_0$ sparse approximation problem with the tool of deep learning, by proposing Deep $\ell_0$ Encoders. Two typical forms, the $\ell_0$ regularized problem and the $M$-sparse problem, are investigated. Based on solid iterative algorithms, we model them as feed-forward neural networks, through introducing novel neurons and pooling functions. Enforcing such structural priors acts as an effective network regularization. The deep encoders also enjoy faster inference, larger learning capacity, and better scalability compared to conventional sparse coding solutions. Furthermore, under task-driven losses, the models can be conveniently optimized from end to end. Numerical results demonstrate the impressive performances of the proposed encoders.
\end{quote}
\end{abstract}

\section{Dedication}
Zhangyang and Qing would like to dedicate the paper to their friend, \textbf{Mr. Yuan Song} (10/09/1984 - 07/13/2015).


\section{Introduction}

Sparse signal approximation has gained popularity over the last decade. The sparse approximation model suggests that a natural signal could be compactly approximated, by only a few atoms out of a properly given dictionary, where the weights associated with the dictionary atoms are called the sparse codes. Proven to be both robust to noise and scalable to high dimensional data, sparse codes are known as powerful features, and benefit a wide range of signal processing applications, such as source coding \cite{1}, denoising \cite{2}, source separation \cite{3}, pattern classification \cite{wright2009robust}, and clustering \cite{cheng2010learning}.

We are particularly interested in the $\ell_0$-based sparse approximation problem, which is the fundamental formulation of sparse coding \cite{donoho2003optimally}. The nonconvex $\ell_0$ problem is intractable and often instead attacked by minimizing surrogate measures, such as the $\ell_1$-norm, which leads to more tractable computational methods. However, it has been both theoretically and practically discovered that solving $\ell_0$ sparse approximation is still preferable in many cases.

More recently, deep learning has attracted great attentions in many feature learning problems \cite{ImageNet}. The advantages of deep learning lie in its composition of multiple non-linear transformations to yield more abstract and descriptive embedding representations. With the aid of gradient descent, it also scales linearly in time and space with the number of train samples.

It has been noticed that sparse approximation and deep learning bear certain connections \cite{LISTA}. Their similar methodology has been lately exploited in \cite{unfold}, \cite{sprechmann2013supervised}, \cite{PAMI2015}. By turning sparse coding models into deep networks, one may expect faster inference, larger learning capacity, and better scalability. The network formulation also facilitates the integration of task-driven optimization.

In this paper, we investigate two typical forms of $\ell_0$-based sparse approximation problems:  the $\ell_0$ regularized problem, and the $M$-sparse problem. Based on solid iterative algorithms \cite{iterative}, we formulate them as feed-forward neural networks \cite{LISTA}, called \textbf{Deep $\ell_0$ Encoders}, through introducing novel neurons and pooling functions. We study their applications in image classification and clustering; in both cases the models are optimized in a task-driven, end-to-end manner. Impressive performances are observed in numerical experiments.

\section{Related Work}
\subsection{$\ell_0$ and $\ell_1$-based Sparse Approximations}
Finding the sparsest, or minimum $\ell_0$-norm, representation of a signal given a dictionary of basis atoms is an important problem in many application domains. Consider a data sample $\mathbf{x} \in R^{m \times 1}$, that is encoded into its sparse code $\mathbf{a} \in R^{p \times 1}$ using a learned dictionary $\mathbf{D}= [\mathbf{d}_1, \mathbf{d}_2,\cdots, \mathbf{d}_p]$, where $ \mathbf{d}_i \in R^{m \times 1}, i=1,2, \cdots, p$ are the learned atoms. The sparse codes are obtained by solving the \textbf{$\ell_0$ regularized problem} ($\lambda$ is a constant):
\begin{equation}
\begin{array}{l}\label{l0r}
\mathbf{a} = \arg \min_{\mathbf{a}} \frac{1}{2}||\mathbf{x - D a}||_F^2 + \lambda ||\mathbf{a}||_0.
\end{array}
\end{equation}
Alternatively, one could explicitly impose constraints on the number of non-zero coefficients of the solution, by solving the \textbf{$M$-sparse problem}:
\begin{equation}
\begin{array}{l}\label{l0M}
\mathbf{a} = \arg \min_{\mathbf{a}} ||\mathbf{x - D a}||_F^2 \qquad s.t. \quad ||\mathbf{a}||_0 \le M
\end{array}
\end{equation}
Unfortunately, these optimization problems are often intractable because there is a combinatorial increase in the number of local minima as the number of the candidate basis vectors increases. One potential remedy is to employ a convex surrogate measure, such as the $\ell_1$-norm,  in place of the $\ell_0$-norm that leads to a more tractable optimization problem. For example, (\ref{l0r}) could be relaxed as:
\begin{equation}
\begin{array}{l}\label{l1r}
\mathbf{a} = \arg \min_{\mathbf{a}} \frac{1}{2}||\mathbf{x - D a}||_F^2 + \lambda ||\mathbf{a}||_1.
\end{array}
\end{equation}
It creates a unimodal optimization problem that can be solved via linear programming techniques. The downside is that we have now introduced a mismatch between the ultimate goal and the objective function \cite{wipf2004}. Under certain conditions, the minimum $\ell_1$-norm solution equals to the minimum $\ell_0$-norm one \cite{donoho2003optimally}. But in practice, the $\ell_1$ approximation is often used way beyond these conditions, and is thus quite heuristic. As a result, we often get a solution which is not exactly minimizing the original $\ell_0$-norm.

That said, $\ell_1$ approximation is found to work practically well for many sparse coding problems. Yet in certain applications, we intend to control the exact number of nonzero elements, such as basis selection \cite{wipf2004}, where $\ell_0$ approximation is indispensable. Beyond that,  $\ell_0$-approximation are desirable for performance concerns in many ways. In compressive sensing literature, empirical evidence \cite{candes2008enhancing} suggested that using an iterative reweighted $\ell_1$ scheme to approximate the $\ell_0$ solution often improved the quality of signal recovery. In image enhancement, it was shown in \cite{Yuan_2015_CVPR} that $\ell_0$ data fidelity was more suitable for reconstructing images corrupted with impulse noise. For the purpose of image smoothening, the authors of \cite{xu2011image} utilized $\ell_0$ gradient minimization to globally control how many non-zero gradients to approximate prominent structures in a structure-sparsity-management manner. Recent work \cite{wang2015clustering} revealed that $\ell_0$ sparse subspace clustering can completely characterize the set of minimal union-of-subspace structure, without additional separation conditions required by its $\ell_1$ counterpart.

\subsection{Network Implementation of $\ell_1$-Approximation}
\begin{figure}[htbp]
\centering
\includegraphics[resolution=320]{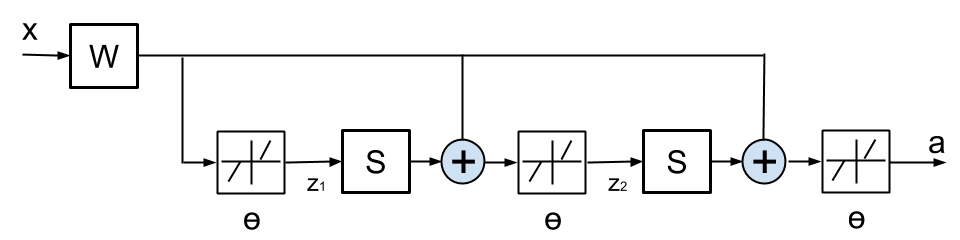}
\caption{A LISTA network \cite{LISTA} with two time-unfolded stages. }
\label{LISTA}
\end{figure}
In \cite{LISTA}, a feed-forward neural network, as illustrated in Fig. \ref{LISTA}, was proposed to efficiently approximate the $\ell_1$-based sparse code $\mathbf{a}$ of the input signal $\mathbf{x}$; the sparse code $a$ is obtained by solving (\ref{l1r}) for a given dictionary $\mathbf{D}$ in advance. The network has a finite number of stages, each of which updates the intermediate sparse code $\mathbf{z}^k$ ($k$ = 1, 2) according to
\begin{equation}
\begin{array}{l}\label{eqnLISTA}
\mathbf{z}^{k+1} = s_{\theta}(\mathbf{Wx} + \mathbf{Sz}^k),
\end{array}
\end{equation}
where $s_{\theta}$ is an element-wise shrinkage function ($\mathbf{u}$ is a vector and $\mathbf{u}_i$ is its $i$-th element, $i = 1, 2, ..., p$):
\begin{equation}
\begin{array}{l}\label{threshold}
[s_{\theta}(\mathbf{u})]_i = \text{sign}(\mathbf{u}_i)(|\mathbf{u}_i| - \theta_i)_{+}.
\end{array}
\end{equation}
The parameterized encoder, named learned ISTA (LISTA), is a natural network implementation of the iterative shrinkage and thresholding algorithm (ISTA). LISTA learned all its parameters $\mathbf{W}$, $\mathbf{S}$ and $\theta$ from training data using a back-propagation algorithm \cite{lecun2012efficient}. In this way, a good approximation of the underlying sparse code can be obtained after a fixed small number of stages.

In \cite{sprechmann2013supervised}, the authors leveraged a similar idea on fast trainable regressors and constructed feed-forward network approximations of the learned sparse models. Such a  process-centric view was later extended in \cite{PAMI2015} to develop a principled process of learned deterministic fixed-complexity pursuits, in lieu of iterative proximal gradient descent algorithms, for structured sparse and robust low rank models. Very recently, \cite{unfold} further summarized the methodology of the problem-level and model-based ``deep unfolding'', and developed new architectures as inference algorithms for both Markov random fields and non-negative matrix factorization. Our work shares the similar spirit with those prior wisdoms, yet studies the unexplored $\ell_0$ problems with further insights obtained.

\section{Deep $\ell_0$ Encoders}

\subsection{Deep $\ell_0$-Regularized Encoder}
To solve the optimization problem in (\ref{l0r}), an iterative hard-thresholding (IHT) algorithm was derived in \cite{iterative}:
\begin{equation}
\begin{array}{l}\label{iterl1r}
\mathbf{a}^{k+1} = h_{\lambda^{0.5}}(\mathbf{a}^k + \mathbf{D^T(x - Da}^k)),
\end{array}
\end{equation}
where $\mathbf{a}^k$ denotes the intermediate result of the $k$-th iteration, and $h_{\theta}$ is an element-wise \textbf{hard thresholding} operator:
\begin{equation} \label{ht}
[h_{\lambda^{0.5}} (\mathbf{u})]_i = \left \{ \begin{array}{ll}
0 &  \quad \textup{if}\quad |\mathbf{u}_i| < \lambda^{{0.5}} \\
\mathbf{u}_i  &  \quad \textup{if} \quad |\mathbf{u}_i| \ge \lambda^{{0.5}} \\
\end{array} \right.
\end{equation}

\begin{figure}[htbp]
\centering
\includegraphics[resolution=250]{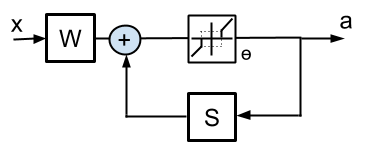}
\caption{The block diagram of solving (\ref{iterl1r}). }.
\label{figISTA}
\end{figure}

Eqn. (\ref{iterl1r}) could be alternatively rewritten as:
\begin{equation}
\begin{array}{l}\label{iterl1r2}
\mathbf{a}^{k+1} = h_\theta(\mathbf{W x} + \mathbf{S a}^k),\\
\mathbf{W} = \mathbf{D}^T, \mathbf{S} = \mathbf{I} - \mathbf{D}^T\mathbf{D}, \theta= \lambda^{{0.5}},
\end{array}
\end{equation}
and expressed as the block diagram in Fig. \ref{figISTA}, which outlines a recurrent network form of solving (\ref{iterl1r}).


\begin{figure*}[tbp]
\centering
\includegraphics[resolution=200]{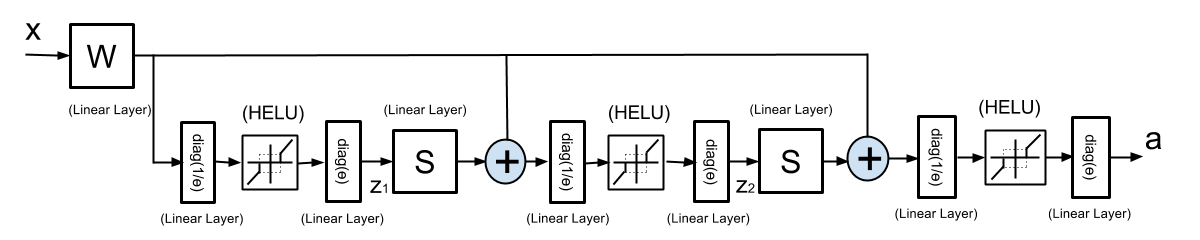}
\caption{Deep $\ell_0$-Regularized Encoder, with two time-unfolded stages.}
\label{figLISTA}
\end{figure*}

\begin{figure*}[tbp]
\centering
\includegraphics[resolution=200]{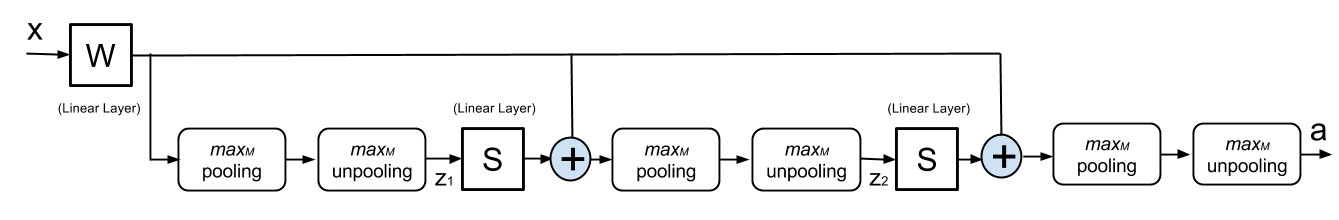}
\caption{Deep $M$-sparse Encoder, with two time-unfolded stages.}
\label{figM}
\end{figure*}

By time-unfolding and truncating Fig. \ref{figISTA} to a fixed number of $K$ iterations ($K$ = 2 in this paper by default)\footnote{We test larger $K$ values (3 or 4). In several cases they do bring performance improvements, but add complexity too..}, we obtain a feed-forward network structure in Fig. \ref{figLISTA}, where  $\mathbf{W}$, $\mathbf{S}$ and $\theta$ are shared among both stages, named \textbf{Deep $\ell_0$-Regularized Encoder}, Furthermore, $\mathbf{W}$,  $\mathbf{S}$ and $\theta$ are all to be learnt, instead of being directly constructed from any pre-computed $\mathbf{D}$. Although the equations in (\ref{iterl1r2}) do not directly apply any more to solving the Deep $\ell_0$-Regularized Encoder, they can usually serve as a high-quality initialization of the latter.

Note that the activation thresholds $\theta$ are less straightforward to update. We rewrite (\ref{threshold}) as: $[h_{\theta}(\mathbf{u})]_i = \theta_i h_1(\mathbf{u}_i/\theta_i)$.  It indicates that the original neuron with trainable thresholds can be decomposed into two linear scaling layers, plus a unit-hard-threshold neuron, the latter of which is called Hard thrEsholding Linear Unit (\textbf{HELU}) by us. The weights of the two scaling layers are diagonal matrices defined by $\theta$ and its element-wise reciprocal, respectively.

\noindent \textbf{Discussion on HELU} While being inspired by LISTA, the differentiating point of Deep $\ell_0$-Regularized Encoder lies in the HELU neuron. Compared to classical neuron functions such as logistic, sigmoid, and ReLU \cite{mhaskar1994choose}, as well as the soft shrinkage and thresholding operation (\ref{threshold}) in LISTA, HELU does not penalize large values, yet enforces strong (in theory infinite) penalty over small values. As such, HELU tends to produce highly sparse solutions.

The neuron form of LISTA (\ref{threshold}) could be viewed as a double-sided and translated variant of ReLU, which is continuous and piecewise linear. In contrast, HELU is a \textbf{discontinuous} function that rarely occurs in existing deep network neurons. As pointed out by \cite{hornik1989multilayer}, HELU has countably many discontinuities and is thus (Borel) measurable, in which case the universal approximation capability of the network is not compromised. However, experiments remind us that the algorithmic learnability with such discontinuous neurons (using popular first-order methods) is in question, and the training is in general hard. For computation concerns, we replace HELU with the following continuous and piecewise linear function $\text{HELU}_{\sigma}$, during network training:
\begin{equation} \label{helu}
[\text{HELU}_{\sigma} (\mathbf{u})]_i = \left \{ \begin{array}{ll}
0 &  \quad \textup{if}\quad |\mathbf{u}_i| \le 1- \sigma \\
\frac{1}{\sigma} (\mathbf{u}_i - 1 + \sigma) &  \quad \textup{if}\quad 1- \sigma < \mathbf{u}_i < 1\\
\frac{1}{\sigma} (\mathbf{u}_i +1 - \sigma) &  \quad \textup{if}\quad - 1 < \mathbf{u}_i < \sigma - 1\\
\mathbf{u}_i  &  \quad \textup{if} \quad |\mathbf{u}_i| \ge 1 \\
\end{array} \right.
\end{equation}
Obviously, $\text{HELU}_{\sigma}$ becomes $\text{HELU}$ when $\sigma \rightarrow$ 0. To approximate HELU, we tend to choose very small $\sigma$, while avoiding putting the training ill-posed. As a practical strategy, we start with a moderate $\sigma$ (0.2 by default), and divide it by 10 after each epoch. After several epoches, $\text{HELU}_{\sigma}$ turns very close to the ideal HELU.

In \cite{rozell2008sparse}, the authors introduced an ideal hard thresholding function for solving sparse coding, whose formulation was close to HELU. Note that \cite{rozell2008sparse} approximates the ideal function with a sigmoid function, which has connections with our $\text{HELU}_{\sigma}$ approximation. In \cite{konda2014zero}, a similar truncated linear ReLU was utilized in the networks.

\subsection{Deep M-Sparse $\ell_0$ Encoder}
Both the $\ell_0$ regularized problem in (\ref{l0r}) and Deep $\ell_0$-Regularized Encoder have no explicit control on the sparsity level of the solution. One would therefore turn to the $M$-sparse problem in (\ref{l0M}), and derive the following iterative algorithm \cite{iterative}:
\begin{equation}
\begin{array}{l}\label{iterl1m}
\mathbf{a}^{k+1} = h_{M}(\mathbf{a}^k + \mathbf{D^T(x - Da}^k)).
\end{array}
\end{equation}
Eqn. (\ref{iterl1m}) resembles (\ref{iterl1r}), except that $h_{M}$ is now a non-linear operator retaining the $M$ coefficients with the \textbf{top $M$-largest absolute values}. Following the same methodology as in the previous section, the iterative form could be time-unfolded and truncated to the \textbf{Deep $M$-sparse Encoder}, as in Fig. \ref{figM}. To deal with the $h_{M}$ operation, we refer to the popular concepts of pooling and unpooling \cite{zeiler2011adaptive} in deep networks, and introduce the pairs of max$_M$ pooling and unpooling, in Fig. \ref{figM}.

\noindent \textbf{Discussion on max$_M$ pooling/unpooling} Pooling is popular in convolutional networks to obtain translation-invariant features \cite{ImageNet}. It is yet less common in other forms of deep networks \cite{gulcehre2014learned}. The unpooling operation was introduced in \cite{zeiler2011adaptive} to insert the pooled values back to the appropriate locations of feature maps for reconstruction purposes.

In our proposed Deep $M$-sparse Encoder, the pooling and unpooling operation pair is used to construct a projection from $R^m$ to its subset $S: \{s \in R^m | ||s||_0 \le M\}$. The max$_M$ pooling and unpooling functions are intuitively defined as:
\begin{equation}
\begin{array}{l}\label{pooling}
[\mathbf{p}_M, \mathbf{idx}_M] = \text{max$_M$.pooling}(\mathbf{u})\\
\mathbf{u}_M = \text{max$_M$.unpooling}(\mathbf{p}_M, \mathbf{idx}_M)
\end{array}
\end{equation}
For each input $\mathbf{u}$, the \textit{pooled map} $\mathbf{p}_M$ records the top $M$-largest values (irrespective of sign), and the \textit{switch} $\mathbf{idx}_M$ records their locations. The corresponding unpooling operation takes the elements in $\mathbf{p}_M$ and places them in $\mathbf{u}_M$ at the locations specified by $\mathbf{idx}_M$, the remaining elements being set to zero. The resulting $\mathbf{u}_M$ is of the same dimension as $\mathbf{u}$ but has exactly no more than $M$ non-zero elements. In back propagation, each position in $\mathbf{idx}_M$ is propagated with the entire error signal.

\subsection{Theoretical Properties}
It is showed in \cite{iterative} that the iterative algorithms in both (\ref{iterl1r}) and (\ref{iterl1m}) are guaranteed not to increase the cost functions. Under mild conditions, their targeted fixed points are local minima of the original problems. As the next step after the time truncation, the deep encoder models are to be solved by the stochastic gradient descent (SGD) algorithm, which converges to stationary points under a few stricter assumptions than ones satisfied in this paper \cite{bottou2010large} \footnote{As a typical case, we use SGD in a setting where it is not guaranteed to converge in theory, but behaves well in practice.}. However, the entanglement of the iterative algorithms and the SGD algorithm makes the overall convergence analysis a serious hardship.

One must emphasize that in each step, the back propagation procedure requires only operations of order O($p$)~\cite{LISTA}. The training algorithm takes O($Cnp$) time ($C$ is the constant absorbing epochs, stage numbers, etc). The testing process is purely feed-forward and is therefore dramatically faster than traditional inference methods by solving (\ref{l0r}) or (\ref{l0M}). SGD is also easy to be parallelized.

\section{Task-Driven Optimization}

It is often desirable to jointly optimize the learned sparse code features and the targeted task so that they mutually reinforce each other. The authors of \cite{jiang2011learning} associated label information with each dictionary item by adding discriminable regularization terms to the objective. Recent work \cite{mairal2012task}, \cite{IJCAI} developed task-driven sparse coding via bi-level optimization models, where ($\ell_1$-based) sparse coding is formulated as the lower-level constraint while a task-oriented cost function is minimized as its upper-level objective. The above approaches in sparse coding are complicated and computationally expensive. It is much more convenient to implement end-to-end task-driven training in deep architectures, by concatenating the proposed deep encoders with certain task-driven loss functions.

In this paper, we mainly discuss two tasks: classification and clustering, while being aware of other immediate extensions, such as semi-supervised learning. Assuming $K$ classes (or clusters), and $\bm{\omega} = [\bm{\omega}_1,..., \bm{\omega}^k]$ as the set of parameters of the loss function, where $\bm{\omega}_i$ corresponds to the $j$-th class (cluster), $j$ = $1, 2, ..., K$. For the \textbf{classification} case, one natural choice is the well-known softmax loss function. For the \textbf{clustering} case, since the true cluster label of each $\mathbf{x}$ is unknown, we define the predicted confidence probability $p_{j}$ that sample $\mathbf{x} $ belongs to cluster $j$, as the likelihood of softmax regression:
\begin{equation}
\begin{array}{l}\label{3}
p_{j}=p(j|\bm{\omega, a})=\frac{e^{- \bm{\omega}_j^T\mathbf{a}}}{\sum_{l=1}^K e^{- \bm{\omega}_l^T\mathbf{a}}}.
\end{array}
\end{equation}
The predicted cluster label of $\mathbf{a}$ is the cluster $j$ where it achieves the largest $p_{j}$.

\section{Experiment}
\subsection{Implementation}
Two proposed deep $\ell_0$ encoders are implemented with the CUDA ConvNet package \cite{ImageNet}. We use a constant learning rate of 0.01 with no momentum, and a batch size of 128. In practice, given that the model is well initialized, the training takes approximately 1 hour on the MNIST dataset, on a workstation with 12 Intel Xeon 2.67GHz CPUs and 1 GTX680 GPU. It is also observed that the training efficiency of our model scales approximately linearly with the size of data.


While many neural networks train well with random initializations without pre-training, given that the training data is sufficient, it has been discovered that poorly initialized networks can hamper the effectiveness of first-order methods (e.g., SGD) \cite{sutskever2013importance}. For the proposed models, it is however much easier to initialize the model in the right regime, benefiting from the analytical relationships between sparse coding and network hyperparameters in (\ref{iterl1r2}).

\subsection{Simulation on $\ell_0$ Sparse Approximation}

We first compare the performance of different methods on $\ell_0$ sparse code approximation. The first 60,000 samples of the MNIST dataset are used for training and the last 10,000 for testing. Each patch is resized to $16 \times 16$ and then preprocessed to remove its mean and normalize its variance. The patches with small standard deviations are discarded. A sparsity coefficient $\lambda$ = 0.5 is used in (\ref{l0r}), and the sparsity level $M$ = 32 is fixed in (\ref{l0M}). The sparse code dimension (dictionary size) $p$ is to be varied.

\begin{table}[htbp]
 \begin{center}
 \caption{Prediction error (\%) comparison of all methods on solving the $\ell_0$-regularized problem (\ref{l0r})}
 \label{l0rerror}
 \begin{tabular}{|c|c|c|c|}
 \hline
Method & $p$ = 128 &  $p$ = 256 &  $p$ = 512   \\
\hline
Iterative (2 iterations) & 17.52 & 18.73 & 22.40 \\
Iterative (5 iterations) & 8.14 & 6.75 & 9.37 \\
Iterative (10 iterations) & 3.55 & 4.33 & 4.08  \\
\hline
Baseline Encoder & 8.94 & 8.76 & 10.17 \\
\hline
Deep $\ell_0$-Regularized Encoder & 0.92 & 0.91 & 0.81  \\
 \hline
 \end{tabular}
 \end{center}
 \end{table}

Our prediction task resembles the setup in \cite{LISTA}: first learning a dictionary from training data, following by solving sparse approximation (\ref{l1r}) with respect to the dictionary, and finally training the network as a regressor from input samples to the solved sparse codes. The only major difference here lies in that unlike the $\ell_1$-based problems, the non-convex $\ell_0$-based minimization could only reach a (non-unique) local minimum. To improve stability, we first solve the $\ell_1$ problems to obtain a good initialization for $\ell_0$ problems, and then run the iterative algorithms (\ref{iterl1r}) or (\ref{iterl1m}) until convergence. The obtained sparse codes are called ``optimal codes'' hereinafter and used in both training and testing evaluation (as ``groundtruth''). One must keep in mind that we are not seeking to produce approximate sparse code for all possible input vectors, but only for \textit{input vectors drawn from the same distribution as our training samples}.

\begin{table}[htbp]
 \begin{center}
 \caption{Prediction error (\%) comparison of all methods on solving the $M$-sparse problem (\ref{l0M})}
 \label{l0Merror}
 \begin{tabular}{|c|c|c|c|}
 \hline
Method & $p$ = 128 &  $p$ = 256 &  $p$ = 512  \\
\hline
Iterative (2 iterations) & 17.23 & 19.27 & 19.31 \\
Iterative (5 iterations) & 10.84 & 12.52 & 12.40 \\
Iterative (10 iterations) & 5.67 & 5.44 & 5.20 \\
\hline
Baseline Encoder & 14.04 & 16.76 & 12.86 \\
\hline
Deep $M$-Sparse Encoder & 2.94 & 2.87 & 3.29 \\
 \hline
 \end{tabular}
 \end{center}
 \end{table}

 \begin{table}[htbp]
 \begin{center}
 \caption{Averaged non-zero support error comparison of all methods on solving the $M$-sparse problem (\ref{l0M})}
 \label{l0Msupport}
 \begin{tabular}{|c|c|c|c|}
 \hline
Method & $p$ = 128 &  $p$ = 256 &  $p$ = 512   \\
\hline
Iterative (2 iterations) & 10.8 & 13.4 & 13.2  \\
Iterative (5 iterations) & 6.1 & 8.0 & 8.8 \\
Iterative (10 iterations) & 4.6 & 5.6 & 5.3 \\
\hline
Deep $M$-Sparse Encoder & 2.2 & 2.7 & 2.7 \\
 \hline
 \end{tabular}
 \end{center}
 \end{table}

We compare the proposed deep $\ell_0$ encoders with the iterative algorithms under different number of iterations. In addition, we include a \textit{baseline encoder} into comparison, which is a fully-connected feed-forward network, consisting of three hidden layers of dimension $p$ with ReLu neurons. The baseline encoder thus has the same parameter capacity as deep $\ell_0$ encoders\footnote{except for the ``diag($\theta$)" layers in Fig. \ref{figLISTA}, each of which contains only $p$ free parameters.}. We apply dropout to the baseline encoders, with the probabilities of retaining the units being 0.9, 0.9, and 0.5. The proposed encoders do not apply dropout.

The deep $\ell_0$ encoders and the baseline encoder are first trained, and all are then evaluated on the testing set. We calculate the total prediction errors, i.e., the normalized squared errors between the optimal codes and the predicted codes, as in Tables \ref{l0rerror} and  \ref{l0Merror}. For the $M$-sparse case, we also compare their recovery of non-zero supports in Table \ref{l0Msupport}, by counting the mismatched nonzero element locations between optimal and predicted codes (averaged on all samples). Immediate conclusions from the numerical results are as follows:
\begin{itemize}
\item The proposed deep encoders have outstanding generalization performances, thanks to the effective regularization brought by their architectures, which are derived from specific problem formulations (i.e., (\ref{l0r}) and (\ref{l0M})) as priors. The ``general-architecture'' baseline encoders, which have the same parameter complexity, appear to overfit the training set and generalize much worse.
\item While the deep encoders only unfold two stages, they outperforms their iterative counterparts even when the later ones have passed 10 iterations. Meanwhile, the former enjoy much faster inference as being feed-forward.
\item The Deep $\ell_0$-Regularized Encoder obtains a particularly low prediction error. It is interpretable that while the iterative algorithm has to work with a fixed $\lambda$, the Deep $\ell_0$-Regularized Encoder is capable of ``fine-tuning'' this hyper-parameter automatically (after diag($\theta$) is initialized from $\lambda$), by exploring the training data structure.
\item The Deep $M$-Sparse Encoder is able to find the nonzero support with high accuracy.
\end{itemize}

\subsection{Applications on Classification}

Since the task-driven models are trained from end to end, \textbf{no pre-computation of  $\mathbf{a}$ is needed}. For classification, we evaluate our methods on the MNIST dataset, and the AVIRIS Indiana Pines hyperspectral image dataset (see \cite{chen2011hyperspectral} for details). We compare our two proposed deep encoders with two competitive sparse coding-based methods: 1) task-driven sparse coding (TDSC) in \cite{mairal2012task}, with the original setting followed and all parameters carefully tuned; 2) a pre-trained LISTA followed by supervised tuning with softmax loss. Note that for Deep $M$-Sparse Encoder, $M$ is not known in advance and has to be tuned. To our surprise, the fine-tuning of $M$ is likely to improve the performances significantly, which is analyzed next. The overall error rates are compared in Tables \ref{MNISTc} and \ref{Pinec}.

 \begin{table}[htbp]
 \begin{center}
 \caption{Classification error rate (\%) comparison of all methods on the MNIST dataset}
 \label{MNISTc}
 \begin{tabular}{|c|c|c|c|}
 \hline
Method & $p$ = 128 &  $p$ = 256 &  $p$ = 512   \\
\hline
TDSC & 0.71 & 0.55 & 0.53 \\
\hline
Tuned LISTA & 0.74 & 0.62 & 0.57 \\
\hline
Deep $\ell_0$-Regularized & 0.72 & 0.58 & 0.52  \\
\hline
Deep $M$-Sparse ($M$ = 10) & 0.72 & 0.57 & 0.53 \\
 \hline
Deep $M$-Sparse ($M$ = 20) & 0.69 & 0.54 & 0.51 \\
 \hline
 Deep $M$-Sparse ($M$ = 30) & 0.73 & 0.57 & 0.52 \\
 \hline
 \end{tabular}
 \end{center}
 \end{table}

  \begin{table}[htbp]
 \begin{center}
 \caption{Classification error rate (\%) comparison of all methods on the AVIRIS Indiana Pines dataset}
 \label{Pinec}
 \begin{tabular}{|c|c|c|c|}
 \hline
Method & $p$ = 128 &  $p$ = 256 &  $p$ = 512   \\
\hline
TDSC & 15.55 & 15.27 & 15.21 \\
\hline
Tuned LISTA & 16.12 & 16.05 & 15.97 \\
\hline
Deep $\ell_0$-Regularized & 15.20 & 15.07 & 15.01  \\
\hline
Deep $M$-Sparse ($M$ = 10) & 13.77 & 13.56 & 13.52 \\
 \hline
Deep $M$-Sparse ($M$ = 20) & 14.67 & 14.23 & 14.07 \\
 \hline
 Deep $M$-Sparse ($M$ = 30) & 15.14 & 15.02 & 15.00 \\
 \hline
 \end{tabular}
 \end{center}
 \end{table}
In general, the proposed deep $\ell_0$ encoders provide superior results to the deep $\ell_1$-based method (tuned LISTA). TDSC also generates competitive results, but at the cost of the high complexity for inference, i.e., solving conventional sparse coding. It is of particular interest to us that when supplied with specific $M$ values, the  Deep $M$-Sparse encoder can generate remarkably improved results  \footnote{To get a good estimate of $M$, one might first try to perform (unsupervised) sparse coding on a subset of samples.}. Especially in Table \ref{Pinec}, when $M$ = 10, the error rate is around 1.5\% lower than that of  $M$ = 30. Note that in the AVIRIS Indiana Pines dataset, the training data volume is much smaller than that of MNIST. In that way, we conjecture that it might not be sufficiently effective to depend the training process fully on data; instead, to craft a stronger sparsity prior by smaller $M$ could help learn more discriminative features\footnote{Interestingly, there are a total of 16 classes in the AVIRIS Indiana Pines dataset When $p$ = 128, each class has on average 8 ``dictionary atoms'' for class-specific representation. Therefore $M$ = 10 approximately coincides with the sparse representation classification (SRC) principle \cite{chen2011hyperspectral} of forcing sparse codes to be compactly focused on one class of atoms.}. Such a behavior provides us with a important hint to \textbf{impose suitable structural priors to deep networks.}

\subsection{Applications on Clustering}

For clustering, we evaluate our methods on the COIL 20 and the CMU PIE dataset \cite{sim2002cmu}. Two state-of-the-art methods to compare are the jointly optimized sparse coding and clustering method proposed in \cite{IJCAI}, as well as the graph-regularized deep clustering method in \cite{ICDM}\footnote{Both papers train their model under both soft-max and max-margin type losses. To ensure fair comparison, we adopt the former, with the same form of loss function as ours.}. The overall error rates are compared in Tables \ref{Coilc} and \ref{CMUc}.

Note that the method in \cite{ICDM} incorporated Laplacian regularization as an additional prior while the others not. It is thus no wonder that this method often performs better than others. Even without any graph information utilized, the proposed deep encoders are able to obtain very close performances, and outperforms \cite{ICDM} in certain cases. On the COIL 20 dataset, the lowest error rate is reached by the Deep $M$-Sparse ($M$ = 10) Encoder, when $p$ = 512, followed by the Deep $\ell_0$-Regularized Encoder.
 \begin{table}[htbp]
 \begin{center}
 \caption{Clustering error rate (\%) comparison of all methods on the COIL 20 dataset}
 \label{Coilc}
 \begin{tabular}{|c|c|c|c|}
 \hline
Method & $p$ = 128 &  $p$ = 256 &  $p$ = 512   \\
\hline
\cite{IJCAI} & 17.75 & 17.14 & 17.15 \\
\hline
\cite{ICDM}  & 14.47 & 14.17 & 14.08 \\
\hline
Deep $\ell_0$-Regularized & 14.52 & 14.27 & 14.06  \\
\hline
Deep $M$-Sparse ($M$ = 10) & 14.59 & 14.25 & 14.03 \\
 \hline
Deep $M$-Sparse ($M$ = 20) & 14.84 & 14.33 & 14.15 \\
 \hline
 Deep $M$-Sparse ($M$ = 30) & 14.77 & 14.37 & 14.12 \\
 \hline
 \end{tabular}
 \end{center}
 \end{table}

  \begin{table}[htbp]
 \begin{center}
 \caption{Clustering error rate (\%) comparison of all methods on the CMU PIE dataset}
 \label{CMUc}
 \begin{tabular}{|c|c|c|c|}
 \hline
Method & $p$ = 128 &  $p$ = 256 &  $p$ = 512   \\
\hline
\cite{IJCAI}  & 17.50 & 17.26 & 17.20 \\
\hline
\cite{ICDM}  & 16.14 & 15.58 & 15.09 \\
\hline
Deep $\ell_0$-Regularized & 16.08 & 15.72 & 15.41  \\
\hline
Deep $M$-Sparse ($M$ = 10) & 16.77 & 16.46 & 16.02 \\
 \hline
Deep $M$-Sparse ($M$ = 20) & 16.44 & 16.23 & 16.05 \\
 \hline
 Deep $M$-Sparse ($M$ = 30) & 16.46 & 16.17 & 16.01 \\
 \hline
 \end{tabular}
 \end{center}
 \end{table}

On the CMU PIE dataset, the Deep $\ell_0$-Regularized Encoder leads to competitive accuracies with \cite{ICDM}, and outperforms all Deep $M$-Sparse Encoders with noticeable margins, which is different from other cases. Pervious work discovered that sparse approximations over CMU PIE had significant errors \cite{yang2010supervised}, which is also verified by us. Therefore, hardcoding exact sparsity could even hamper the model performance.

\noindent \textbf{Remark: }
From those experiments, we gain additional insights in designing deep architectures:
\begin{itemize}
\item If one expects the model to explore the data structure by itself, and provided that there is sufficient training data, then the Deep $\ell_0$-Regularized Encoder (and its peers)  might be preferred as its all parameters, including the desired sparsity, are fully learnable from the data.
\item If one has certain correct prior knowledge of the data structure, including but not limited to the exact sparsity level, one should choose Deep $M$-Sparse Encoder, or other models of its type that are designed to maximally enforce that prior. The methodology could be especially useful when the training data is less than sufficient.
\end{itemize}
We hope the above insights could be of reference to many other deep learning models.

\section{Conclusion}

We propose Deep $\ell_0$ Encoders to solve the $\ell_0$ sparse approximation problem. Rooted  in solid iterative algorithms, the deep $\ell_0$ regularized encoder and deep $M$-sparse encoder are developed, each designed to solve one typical formulation, accompanied with the introduction of the novel HELU neuron and max$_M$ pooling/unpooling. When applied to specific tasks of classification and clustering, the models are optimized in an end-to-end manner. The latest deep learning tools enable us to solve them in a highly effective and efficient fashion. They not only provide us with impressive performances in numerical experiments, but also inspire us with important insights into designing deep models.

\bibliographystyle{aaai}
\bibliography{aaai}

\begin{thebibliography}{}

\bibitem[\protect\citeauthoryear{Blumensath and Davies}{2008}]{iterative}
Blumensath, T., and Davies, M.~E.
\newblock 2008.
\newblock Iterative thresholding for sparse approximations.
\newblock {\em Journal of Fourier Analysis and Applications} 14(5-6):629--654.

\bibitem[\protect\citeauthoryear{Bottou}{2010}]{bottou2010large}
Bottou, L.
\newblock 2010.
\newblock Large-scale machine learning with stochastic gradient descent.
\newblock In {\em Proceedings of COMPSTAT'2010}. Springer.
\newblock  177--186.

\bibitem[\protect\citeauthoryear{Candes, Wakin, and
  Boyd}{2008}]{candes2008enhancing}
Candes, E.~J.; Wakin, M.~B.; and Boyd, S.~P.
\newblock 2008.
\newblock Enhancing sparsity by reweighted l1 minimization.
\newblock {\em Journal of Fourier analysis and applications} 14(5-6):877--905.

\bibitem[\protect\citeauthoryear{Cheng \bgroup et al\mbox.\egroup
  }{2010}]{cheng2010learning}
Cheng, B.; Yang, J.; Yan, S.; Fu, Y.; and Huang, T.~S.
\newblock 2010.
\newblock Learning with l1 graph for image analysis.
\newblock {\em TIP} 19(4).

\bibitem[\protect\citeauthoryear{Davies and Mitianoudis}{2004}]{3}
Davies, M., and Mitianoudis, N.
\newblock 2004.
\newblock Simple mixture model for sparse overcomplete ica.
\newblock {\em IEE Proceedings-Vision, Image and Signal Processing}
  151(1):35--43.

\bibitem[\protect\citeauthoryear{Donoho and Elad}{2003}]{donoho2003optimally}
Donoho, D.~L., and Elad, M.
\newblock 2003.
\newblock Optimally sparse representation in general (nonorthogonal)
  dictionaries via l1 minimization.
\newblock {\em Proceedings of the National Academy of Sciences}
  100(5):2197--2202.

\bibitem[\protect\citeauthoryear{Donoho \bgroup et al\mbox.\egroup }{1998}]{1}
Donoho, D.~L.; Vetterli, M.; DeVore, R.~A.; and Daubechies, I.
\newblock 1998.
\newblock Data compression and harmonic analysis.
\newblock {\em Information Theory, IEEE Transactions on} 44(6):2435--2476.

\bibitem[\protect\citeauthoryear{Donoho}{1995}]{2}
Donoho, D.~L.
\newblock 1995.
\newblock De-noising by soft-thresholding.
\newblock {\em Information Theory, IEEE Transactions on} 41(3):613--627.

\bibitem[\protect\citeauthoryear{Gregor and LeCun}{2010}]{LISTA}
Gregor, K., and LeCun, Y.
\newblock 2010.
\newblock Learning fast approximations of sparse coding.
\newblock In {\em ICML},  399--406.

\bibitem[\protect\citeauthoryear{Gulcehre \bgroup et al\mbox.\egroup
  }{2014}]{gulcehre2014learned}
Gulcehre, C.; Cho, K.; Pascanu, R.; and Bengio, Y.
\newblock 2014.
\newblock Learned-norm pooling for deep feedforward and recurrent neural
  networks.
\newblock In {\em Machine Learning and Knowledge Discovery in Databases}.
  Springer.
\newblock  530--546.

\bibitem[\protect\citeauthoryear{Hershey, Roux, and Weninger}{2014}]{unfold}
Hershey, J.~R.; Roux, J.~L.; and Weninger, F.
\newblock 2014.
\newblock Deep unfolding: Model-based inspiration of novel deep architectures.
\newblock {\em arXiv preprint arXiv:1409.2574}.

\bibitem[\protect\citeauthoryear{Hornik, Stinchcombe, and
  White}{1989}]{hornik1989multilayer}
Hornik, K.; Stinchcombe, M.; and White, H.
\newblock 1989.
\newblock Multilayer feedforward networks are universal approximators.
\newblock {\em Neural networks} 2(5):359--366.

\bibitem[\protect\citeauthoryear{Jiang, Lin, and
  Davis}{2011}]{jiang2011learning}
Jiang, Z.; Lin, Z.; and Davis, L.~S.
\newblock 2011.
\newblock Learning a discriminative dictionary for sparse coding via label
  consistent k-svd.
\newblock In {\em CVPR},  1697--1704.
\newblock IEEE.

\bibitem[\protect\citeauthoryear{Konda, Memisevic, and
  Krueger}{2014}]{konda2014zero}
Konda, K.; Memisevic, R.; and Krueger, D.
\newblock 2014.
\newblock Zero-bias autoencoders and the benefits of co-adapting features.
\newblock {\em arXiv preprint arXiv:1402.3337}.

\bibitem[\protect\citeauthoryear{Krizhevsky, Sutskever, and
  Hinton}{2012}]{ImageNet}
Krizhevsky, A.; Sutskever, I.; and Hinton, G.~E.
\newblock 2012.
\newblock Imagenet classification with deep convolutional neural networks.
\newblock In {\em NIPS},  1097--1105.

\bibitem[\protect\citeauthoryear{LeCun \bgroup et al\mbox.\egroup
  }{2012}]{lecun2012efficient}
LeCun, Y.~A.; Bottou, L.; Orr, G.~B.; and M{\"u}ller, K.-R.
\newblock 2012.
\newblock Efficient backprop.
\newblock In {\em Neural networks: Tricks of the trade}. Springer.
\newblock  9--48.

\bibitem[\protect\citeauthoryear{Mairal, Bach, and
  Ponce}{2012}]{mairal2012task}
Mairal, J.; Bach, F.; and Ponce, J.
\newblock 2012.
\newblock Task-driven dictionary learning.
\newblock {\em TPAMI} 34(4):791--804.

\bibitem[\protect\citeauthoryear{Mhaskar and
  Micchelli}{1994}]{mhaskar1994choose}
Mhaskar, H.~N., and Micchelli, C.~A.
\newblock 1994.
\newblock How to choose an activation function.
\newblock In {\em NIPS},  319--326.

\bibitem[\protect\citeauthoryear{Rozell \bgroup et al\mbox.\egroup
  }{2008}]{rozell2008sparse}
Rozell, C.~J.; Johnson, D.~H.; Baraniuk, R.~G.; and Olshausen, B.~A.
\newblock 2008.
\newblock Sparse coding via thresholding and local competition in neural
  circuits.
\newblock {\em Neural computation} 20(10):2526--2563.

\bibitem[\protect\citeauthoryear{Sim, Baker, and Bsat}{2002}]{sim2002cmu}
Sim, T.; Baker, S.; and Bsat, M.
\newblock 2002.
\newblock The cmu pose, illumination, and expression (pie) database.
\newblock In {\em Automatic Face and Gesture Recognition, 2002. Proceedings.
  Fifth IEEE International Conference on},  46--51.
\newblock IEEE.

\bibitem[\protect\citeauthoryear{Sprechmann \bgroup et al\mbox.\egroup
  }{2013}]{sprechmann2013supervised}
Sprechmann, P.; Litman, R.; Yakar, T.~B.; Bronstein, A.~M.; and Sapiro, G.
\newblock 2013.
\newblock Supervised sparse analysis and synthesis operators.
\newblock In {\em NIPS},  908--916.

\bibitem[\protect\citeauthoryear{Sprechmann, Bronstein, and
  Sapiro}{2015}]{PAMI2015}
Sprechmann, P.; Bronstein, A.; and Sapiro, G.
\newblock 2015.
\newblock Learning efficient sparse and low rank models.
\newblock {\em TPAMI}.

\bibitem[\protect\citeauthoryear{Sutskever \bgroup et al\mbox.\egroup
  }{2013}]{sutskever2013importance}
Sutskever, I.; Martens, J.; Dahl, G.; and Hinton, G.
\newblock 2013.
\newblock On the importance of initialization and momentum in deep learning.
\newblock In {\em ICML},  1139--1147.

\bibitem[\protect\citeauthoryear{Wang \bgroup et al\mbox.\egroup
  }{2015a}]{IJCAI}
Wang, Z.; Yang, Y.; Chang, S.; Li, J.; Fong, S.; and Huang, T.~S.
\newblock 2015a.
\newblock A joint optimization framework of sparse coding and discriminative
  clustering.
\newblock In {\em IJCAI}.

\bibitem[\protect\citeauthoryear{Wang \bgroup et al\mbox.\egroup
  }{2015b}]{ICDM}
Wang, Z.; Chang, S.; Zhou, J.; and Huang, T.~S.
\newblock 2015b.
\newblock Learning a task-specific deep architecture for clustering.

\bibitem[\protect\citeauthoryear{Wang, Nasrabadi, and
  Huang}{2015}]{chen2011hyperspectral}
Wang, Z.; Nasrabadi, N.~M.; and Huang, T.~S.
\newblock 2015.
\newblock Semisupervised hyperspectral classification using task-driven
  dictionary learning with laplacian regularization.
\newblock {\em TGRS} 53(3):1161--1173.

\bibitem[\protect\citeauthoryear{Wang, Wang, and
  Singh}{2015}]{wang2015clustering}
Wang, Y.; Wang, Y.-X.; and Singh, A.
\newblock 2015.
\newblock Clustering consistent sparse subspace clustering.
\newblock {\em arXiv preprint arXiv:1504.01046}.

\bibitem[\protect\citeauthoryear{Wipf and Rao}{2004}]{wipf2004}
Wipf, D.~P., and Rao, B.~D.
\newblock 2004.
\newblock l0-norm minimization for basis selection.
\newblock In {\em NIPS},  1513--1520.

\bibitem[\protect\citeauthoryear{Wright \bgroup et al\mbox.\egroup
  }{2009}]{wright2009robust}
Wright, J.; Yang, A.~Y.; Ganesh, A.; Sastry, S.~S.; and Ma, Y.
\newblock 2009.
\newblock Robust face recognition via sparse representation.
\newblock {\em TPAMI} 31(2):210--227.

\bibitem[\protect\citeauthoryear{Xu \bgroup et al\mbox.\egroup
  }{2011}]{xu2011image}
Xu, L.; Lu, C.; Xu, Y.; and Jia, J.
\newblock 2011.
\newblock Image smoothing via l 0 gradient minimization.
\newblock In {\em TOG}, volume~30,  174.
\newblock ACM.

\bibitem[\protect\citeauthoryear{Yang, Yu, and
  Huang}{2010}]{yang2010supervised}
Yang, J.; Yu, K.; and Huang, T.
\newblock 2010.
\newblock Supervised translation-invariant sparse coding.
\newblock In {\em Computer Vision and Pattern Recognition (CVPR), 2010 IEEE
  Conference on},  3517--3524.
\newblock IEEE.

\bibitem[\protect\citeauthoryear{Yuan and Ghanem}{2015}]{Yuan_2015_CVPR}
Yuan, G., and Ghanem, B.
\newblock 2015.
\newblock L0tv: A new method for image restoration in the presence of impulse
  noise.

\bibitem[\protect\citeauthoryear{Zeiler, Taylor, and
  Fergus}{2011}]{zeiler2011adaptive}
Zeiler, M.~D.; Taylor, G.~W.; and Fergus, R.
\newblock 2011.
\newblock Adaptive deconvolutional networks for mid and high level feature
  learning.
\newblock In {\em ICCV},  2018--2025.
\newblock IEEE.

\end{thebibliography}

\end{document}